\def\arxivversion{1}
\documentclass[11pt]{article}

\providecommand{\arxivversion}{0}
\newif\ifarxiv
\ifnum\arxivversion=1 \arxivtrue\fi
\ifarxiv\usepackage[preprint]{acl}\else\usepackage[review]{acl}\fi

\usepackage{times}
\usepackage{latexsym}
\usepackage[T1]{fontenc}
\usepackage[utf8]{inputenc}
\usepackage{microtype}
\usepackage{inconsolata}
\usepackage{graphicx}
\usepackage{booktabs}
\usepackage{multirow}
\usepackage{xcolor}
\usepackage{amsmath}
\usepackage{placeins}

\newcommand{\cond}[1]{\textsc{#1}}
\newcommand{\corr}[1]{\texttt{#1}}


\newcommand{\premGapsExcl}{7}
\newcommand{\premGapsAll}{80}
\newcommand{\premGapMean}{$-0.8$}
\newcommand{\scBigNoun}{7.5}
\newcommand{\scBigHyp}{18.1}

\newcommand{\scBigShuf}{10.2}
\newcommand{\scSmallNoun}{4.0}
\newcommand{\scSmallHyp}{17.0}

\newcommand{\scSmallShuf}{6.7}

\title{Shortcuts or Memorisation? Revisiting Data Corruption in Natural Language Understanding with Large Language Models}

\author{Aarne Talman \\
  University of Helsinki \\
  \texttt{aarne.talman@helsinki.fi} \And
  Stergios Chatzikyriakidis \\
  Institute of Language and Speech Processing \\
  Athena Research Centre, Athens \\
  University of Turku \\
  \texttt{s.chatzikyriakidis@athenarc.gr}}

\begin{document}
\maketitle

\begin{abstract}
Pre-trained encoders fine-tuned on natural language inference (NLI) and other language understanding tasks keep much of their accuracy after the input is perturbed, e.g. by removing all nouns or shuffling the words.
Does the same hold for large language models (LLMs), and if so, is it because they have seen the benchmarks during pre-training?
In this paper we revisit these perturbation experiments with eight LLMs alongside fine-tuned encoders on GLUE and the three rounds of ANLI.
Regardless of model size, every system that answers directly retains 36--72\% of its above-chance performance without nouns on MNLI compared to only 14--29\% on ANLI.
Only a model that reasons before answering departs from this pattern.
We also check contamination indicators (including exact $n$-gram counts in the public pre-training data of two of the models) and find no link between exposure and robustness.
Our strongest evidence comes from 368 new, human-validated items that none of the models can have seen.
All eight LLMs keep more of their performance without nouns on items written in MNLI's style than on items written without its cues.
When the perturbation instead removes the function words carrying such cues, this advantage reverses.
In light of these results, we argue that robustness to perturbation on familiar NLI benchmarks reflects dataset shortcuts rather than memorisation, and tells us more about the benchmark than about the model.
\end{abstract}

\section{Introduction}
Benchmark performance remains the primary evidence supporting claims regarding the language understanding capabilities of current models.
However, achieving high scores on benchmarks does not mean that the model solves the problem as intended by the benchmark.
One method to verify this is to transform the input data by removing information that humans need, and then check if the model still makes accurate predictions.
While previous studies referred to this as data corruption, we use \emph{word-class ablation} for word removal and \emph{word-order perturbation} for changing word order.
\citet{talman-etal-2021-nli} and \citet{talman-etal-2022-data} performed this type of evaluation on fine-tuned encoders.
Accuracy was largely maintained on MNLI and various GLUE tasks even when nouns, verbs, or other parts of speech were all removed or when the word order was shuffled. This was the case even though human readers found many of the resulting items difficult or impossible to label.

For large language models (LLMs), the same questions are raised again in different settings.
LLMs are pre-trained on web-scale corpora that may already contain benchmark items, and are increasingly being evaluated without task-specific fine-tuning.
If an LLM maintains accuracy on perturbed MNLI items, two explanations are possible.
The first is memorisation: the model recognises the item even after perturbation because it encountered it during pre-training.
The second is that the item can be solved from superficial cues that remain after perturbation, such as negation words in the hypothesis (shortcuts or annotation artefacts).
These two explanations lead to different empirical predictions.
According to the memorisation account, items present in the pre-training corpus should be more robust than absent items, and the effect should be greatest on the benchmarks with the highest potential for leakage. Conversely, if robustness stems from shortcuts, items that a model can solve solely from the hypothesis should be more robust regardless of prior exposure, and this advantage should disappear if the cue-providing words themselves are removed.

This paper tests both of these predictions.
First, we reproduce previously published encoder results on the MNLI dataset within 1 percentage point, and then re-examine the perturbation experiments of \citet{talman-etal-2021-nli,talman-etal-2022-data} using four open-weight LLMs, four hosted LLMs, and fine-tuned encoders.
The main contributions of this paper are as follows:
\begin{itemize}
\item We demonstrate that the gap between MNLI and ANLI lies in the dataset, not the model. Regardless of model size, all systems that answer directly keep much more of their performance on MNLI than on ANLI when nouns are removed (36--72\% vs.\ 14--29\%). Only the model that reasons before answering deviates from this pattern.
\item Introducing perturbations to the inputs causes the LLMs to be biased toward a single label, so that raw accuracy can underestimate how much performance is retained, and we measure robustness after correcting for these biases.
\item We find no association between the three contamination indicators and robustness. Rather, in all three of the most robust LLMs on MNLI, items solvable solely from the hypothesis are more robust, and this advantage is reversed when function words are removed.
\item We construct 368 new, human-validated NLI items written either in the MNLI style or without cues. All eight LLMs exhibit the same shortcut pattern in these unseen items.
\item We compare MNLI with BioASQ, a benchmark without hypothesis cues. BioASQ is also robust to perturbation, but for a different reason: many of its questions can be answered directly from biomedical knowledge.
\end{itemize}
In other words, robustness against perturbation alone cannot prove language understanding or memorisation.
What it reveals depends on which cues remain intact even after perturbation is applied.

\ifarxiv Code and aggregate results are available at \url{https://github.com/aarnetalman/shortcuts-or-memorisation}; the fresh items will be released on publication.\else Code and aggregate results will be released on publication; the fresh items will be released at the same time.\fi

\section{Related Work}

\paragraph{Perturbed input.}
Several studies have shown that models fine-tuned on NLI and other GLUE tasks are remarkably insensitive to changes in input form.
\citet{sinha-etal-2021-unnatural} show that NLI models predict the same label across many permutations of a sentence pair, while \citet{pham-etal-2021-order} and \citet{gupta-etal-2021-bert} report comparable findings for word order on GLUE.
Even at the pre-training stage, \citet{sinha-etal-2021-masked} find that training on shuffled text causes very little degradation on downstream tasks.
For word removal, \citet{talman-etal-2021-nli} ablate entire word classes in MNLI and ANLI and find that models trained on the ablated data still reach high accuracy.
\citet{talman-etal-2022-data} extend these experiments across GLUE and show that the outcome varies considerably by task.
Removing nouns and verbs has the largest impact, whereas some tasks barely change at all.
Nearly all of this work, however, predates instruction-tuned LLMs, which are normally evaluated without task-specific training.

\paragraph{Shortcuts in NLI.}
Crowdsourced NLI corpora are known to contain annotation artefacts that allow a classifier to predict the label without looking at the premise.
\citet{gururangan-etal-2018-annotation} and \citet{poliak-etal-2018-hypothesis} demonstrate that hypothesis-only models perform well above chance on both SNLI and MNLI, with negation and other function words serving as strong indicators of contradiction.
In addition, \citet{mccoy-etal-2019-right} show that NLI models rely heavily on lexical overlap heuristics.
ANLI \citep{nie-etal-2020-adversarial} was constructed adversarially to address this issue, asking annotators to write hypotheses that fool strong models and thereby reducing the scope for such cues.

\paragraph{Contamination.}
As pre-training corpora grow in scale, evaluation benchmarks are increasingly likely to be included in them \citep{sainz-etal-2023-nlp}.
Where the pre-training data are accessible, benchmark overlap can be measured directly, e.g. through exact $n$-gram counts \citep{liu-etal-2024-infinigram}.
When the training data are not open, contamination has to be probed indirectly through model behaviour.
Existing approaches either ask the model to complete or fill in parts of benchmark items \citep{golchin-surdeanu-2024-time,deng-etal-2024-investigating}, or inspect the likelihood assigned to an item's least probable tokens \citep{shi-etal-2024-detecting,zhang-etal-2025-mink}.

\paragraph{Answer bias.}
When LLMs are prompted to output a class label, they often favour particular answers regardless of the input, although recentring the answer scores removes this preference \citep{zhao-etal-2021-calibrate,holtzman-etal-2021-surface}.

Our work connects these threads: on a shared set of models and items, we test whether robustness under perturbation follows pre-training contamination or dataset shortcuts, and we pull the two apart using newly written items that none of the models can have seen.

\section{Experimental Set-up}
\label{sec:method}

\subsection{Perturbations}
We use the perturbations of \citet{talman-etal-2022-data}, applied independently to each text field of an item, and keep the earlier condition names (\cond{Corrupt-Train}, \cond{Corrupt-Test}).
Words are tagged with the universal part-of-speech tagset \citep{petrov-etal-2012-universal} using NLTK \citep{bird-etal-2009-nltk}.
\corr{drop-X} removes all words of class X (e.g.\ \corr{drop-NOUN}); \corr{keep-X+Y} keeps only words of the listed classes (e.g.\ \corr{keep-NOUN+VERB}, which removes all function words); \corr{shuffle-n$k$} splits each field into consecutive $k$-grams and shuffles their order \citep{pham-etal-2021-order}; \corr{hyp-only} removes the premise.
Perturbed text is detokenised and fields that become empty are kept, as in \citet{talman-etal-2022-data}; for the reproduction of \citet{talman-etal-2021-nli} we use that paper's conventions and, where available, its released files.
In total there are 19 perturbations; for the hosted models we use a core set of ten.\footnote{\corr{clean}, \corr{drop-NOUN}, \corr{drop-VERB}, \corr{drop-ADJ}, \corr{drop-DET}, \corr{drop-PRON}, \corr{shuffle-n1}, \corr{shuffle-n3}, \corr{keep-NOUN+VERB}, \corr{hyp-only}.}

\subsection{Data}
We evaluate on the development sets of the GLUE tasks \citep{wang-etal-2018-glue}, using the matched development set for MNLI \citep{williams-etal-2018-broad}, and on the test sets of the three ANLI rounds \citep{nie-etal-2020-adversarial}.
The original splits and metrics are kept: accuracy for MNLI, QNLI, RTE, SST-2 and ANLI, F1 for MRPC and QQP, Matthews correlation for CoLA and Pearson correlation for STS-B.
For the hosted models, for five-shot prompting and for the contamination checks we use a fixed label-stratified sample of 2,000 items for MNLI and QQP.

\paragraph{Fresh items.}
As a control that no model can have seen, we built 368 new NLI items (Appendix~\ref{app:fresh}).
Premises are sentences from Wikipedia articles created, and from Wikipedia's daily current-events pages written, after 1 September 2026, excluding any sentence with a 13-gram found in OLMo's pre-training data.
Two models that we do not evaluate, Llama 4 Maverick and GPT-5.5, wrote one hypothesis per label for each premise, in two styles: following MNLI's annotation instructions with MNLI training pairs as examples (\emph{MNLI style}), and with instructions that rule out the best-known hypothesis-only cues such as negation in contradictions (\emph{cue-free}).
A further set pairs sentences from the first and the current version of new Wikipedia articles.
Fourteen volunteers labelled the items without seeing the intended label; we keep items whose labels all agree with the intended one and that nobody flagged, giving 168 MNLI-style, 148 cue-free and 52 naturally occurring pairs.

\subsection{Models}

\paragraph{Fine-tuned encoders.}
As a reference we fine-tune RoBERTa-base \citep{liu-etal-2019-roberta}, BERT-base \citep{devlin-etal-2019-bert}, ModernBERT-base \citep{warner-etal-2025-smarter} and XLM-R-base \citep{conneau-etal-2020-unsupervised} on clean training data (3 epochs, batch size 32, learning rate $2\cdot10^{-5}$, three seeds).
Ablated training data (\cond{Corrupt-Train}) is used only for the reproduction.
For ANLI we use the released RoBERTa-large model of \citet{nie-etal-2020-adversarial}, trained on SNLI, MNLI, FEVER-NLI and ANLI.

\paragraph{Open LLMs.}
We use Qwen3.8-27B, Muse Glimmer 30B and the base versions of OLMo 3 7B and 32B, whose pre-training data are public.\footnote{Hugging Face checkpoints \texttt{Qwen/Qwen3.8-27B}, \texttt{meta-models/Muse-Glimmer-30B}, \texttt{allenai/Olmo-3-1025-7B} and \texttt{allenai/Olmo-3-1125-32B}.}
Each item is formatted with a task instruction that ends in a one-word answer choice (e.g.\ \emph{Answer with one word: entailment, neutral, or contradiction.}), with the model's chat template where it has one and reasoning switched off.
The prediction is the answer word with the highest log-probability.
In the five-shot setting the prompt contains five training examples, either clean or with nouns removed (\corr{drop-NOUN}), the in-context analogue of \cond{Corrupt-Train}.

\paragraph{Hosted models.}
Mistral Large 3, DeepSeek-V4-Pro and GPT-4o\footnote{Azure AI Foundry deployments \texttt{Mistral-Large-3}, \texttt{DeepSeek-V4-Pro}, \texttt{gpt-4o} and \texttt{grok-4.7}, queried in November 2026.} are prompted with the same instructions at temperature 0, with reasoning switched off.
For GPT-4o and DeepSeek-V4-Pro we also record the log-probabilities of the top answer tokens.
Grok 4.7, which cannot be run without reasoning, is evaluated on a subset of 400 items per task under five perturbations.

\subsection{Measures}
\label{sec:measures}
Raw accuracy conflates two things: losing the information needed for the task, and shifting towards one answer.
We therefore also report \emph{centred accuracy}: for each perturbation, the mean score of each answer over all items is subtracted before taking the argmax, which removes a constant preference without using gold labels \citep[cf.][]{zhao-etal-2021-calibrate}.
To compare systems with different clean performance we report the share of above-chance performance a system retains,
\begin{equation*}
R = 100 \cdot \frac{s_{\text{pert}} - c}{s_{\text{clean}} - c},
\end{equation*}
where $c$ is chance accuracy ($1/k$ for $k$ labels).
$R=100$ means no loss and $R=0$ means a drop to chance.
Unless stated otherwise, $R$ is computed from centred accuracy for the open LLMs, and from accuracy for the encoders and for hosted models without log-probabilities.
Confidence intervals are 95\% bootstrap intervals over items (1,000 resamples, or 500 for the group comparisons in Section~\ref{sec:contamination}).

\section{Results}
\label{sec:results}

Before moving on to LLMs, we verify that our pipeline reproduces the previous encoder results (Appendix~\ref{app:anli}). On MNLI, all 8 reproduced scores for RoBERTa-base and BERT-base are within 0.6 points of the published ones, the other GLUE tasks are within about 3 points with two explained exceptions, and the public ANLI model of \citet{nie-etal-2020-adversarial} is within 3 points on all three rounds, clean and perturbed.

\subsection{Perturbation Effects Across Systems}

\begin{table*}[t]
\centering
\small
\begin{tabular}{lccccccccc}
\toprule
& \multicolumn{3}{c}{MNLI} & \multicolumn{3}{c}{RTE} & \multicolumn{3}{c}{ANLI (R1--R3 average)} \\
\cmidrule(lr){2-4}\cmidrule(lr){5-7}\cmidrule(lr){8-10}
System & Clean & $R$ noun & $R$ verb & Clean & $R$ noun & $R$ verb & Clean & $R$ noun & $R$ verb \\
\midrule
RoBERTa (fine-tuned)$^\ddagger$ & 87.8 & 72 & 81 & 70.8 & 19 & 85 & 56.1 & 24 & 66 \\
\midrule
Qwen3.8-27B & 88.9 & 63 & 74 & 91.3 & 40 & 73 & 75.2 & 29 & 60 \\
OLMo 3 32B & 89.3 & 70 & 79 & 90.6 & 30 & 76 & 61.6 & 19 & 49 \\
OLMo 3 7B & 88.8 & 68 & 76 & 87.0 & 40 & 66 & 51.4 & 14 & 50 \\
Muse Glimmer 30B & 67.2 & 47 & 66 & 80.9 & 29 & 76 & 55.6 & 25 & 56 \\
\midrule
GPT-4o & 81.5 & 56 & 71 & 80.1 & 48 & 75 & 71.4 & 26 & 64 \\
DeepSeek-V4-Pro & 74.2 & 49 & 64 & 88.5 & 37 & 73 & 57.7 & 25 & 58 \\
Mistral Large 3$^\dagger$ & 80.7 & 36 & 50 & 85.6 & 17 & 63 & 63.6 & 23 & 47 \\
\midrule
Grok 4.7 (reasoning)$^{\dagger\S}$ & 71.8 & 43 & 60 & 88.8 & 34 & 87 & 77.9 & 38 & 68 \\
 
\bottomrule
\end{tabular}
\caption{Clean accuracy (\%) and share of performance above chance ($R$, \%) maintained under the \corr{drop-NOUN} and \corr{drop-VERB} conditions. $R$ is calculated from centred accuracy for open-weight LLMs and from accuracy for others (Section~\ref{sec:measures}). $^\ddagger$For MNLI and RTE, the RoBERTa-base model fine-tuned on clean data is used, and for ANLI, the public RoBERTa-large model from \citet{nie-etal-2020-adversarial} is used. $^\dagger$Answer scores are unavailable, and MNLI is evaluated on a 2,000-item sample. $^\S$Reasoning model, first 400 items per task.}
\label{tab:main}
\end{table*}

\begin{figure*}[t]
\centering
\includegraphics[width=\textwidth]{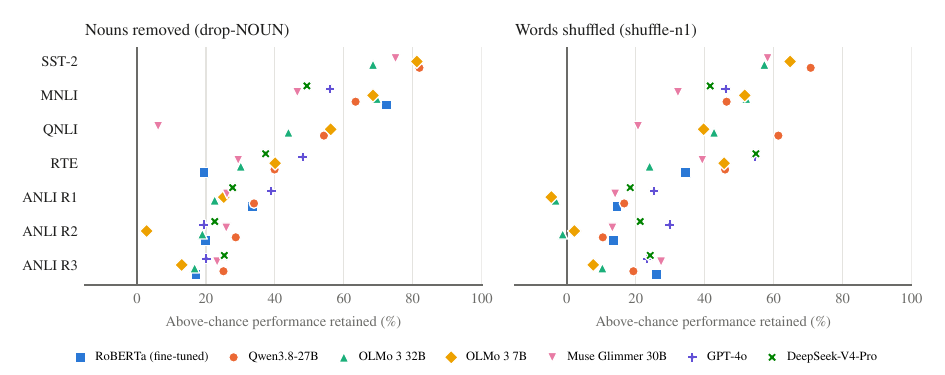}
\caption{Share of performance above chance maintained after noun removal (left) and word shuffling (right). Open-weight LLMs are scored zero-shot by centred accuracy, and RoBERTa is scored by accuracy (RoBERTa-large for ANLI, not run with shuffling). Values less than 0 indicate performance below chance.}
\label{fig:retained}
\end{figure*}

Table~\ref{tab:main} and Figure~\ref{fig:retained} summarise how much performance the systems maintain. An LLM not trained on the task at all is expected to behave differently from a fine-tuned encoder. In some respects, that is the case. Excluding nouns, open-weight LLMs maintain slightly lower MNLI performance than RoBERTa (47--70\% vs. 72\%), but maintain higher RTE performance (29--40\% vs. 19\%).

However, the more notable result is what the systems share in common. Excluding nouns, open-weight LLMs retain 47--70\% of their above-chance performance on MNLI, whereas on ANLI they retain only 14--29\% (averaged over the three rounds). RoBERTa shows the same contrast (72\% vs. 24\%), and similar results are observed in the hosted models GPT-4o (56\% vs. 26\%), DeepSeek-V4-Pro (49\% vs. 25\%), and Mistral Large 3 (36\% vs. 23\%), which is scored on raw answers. When word order is shuffled, the performance difference becomes even more pronounced. Open-weight LLMs maintain 32--52\% on MNLI and up to 27\% on ANLI rounds, whereas OLMo 3 is close to or below chance on ANLI R1 and R2 (Figure~\ref{fig:retained}). Among the GLUE tasks, CoLA, where labels are determined by word order and morphology, is most affected, whereas SST-2 maintains most of its performance, similar to \citet{talman-etal-2022-data}.

Model size does not substantially affect robustness. OLMo 3 7B and 32B were trained on the same data, and the larger model shows 10 points higher performance on clean ANLI. However, both models maintain similar retention on MNLI without nouns (68\% and 70\%), while the larger model shows only slightly higher retention on ANLI (19\% vs. 14\%). Unlike training an encoder on ablated data, five ablated in-context examples also do not help (Appendix~\ref{app:fewshot}).

\subsection{Perturbed Input Shifts the Answer Distribution}
\label{sec:answer-bias}

Perturbed input not only removes information but also biases LLMs toward a single answer. Table~\ref{tab:answer-bias} in Appendix~\ref{app:full} shows the proportion of items receiving the most frequent answer. When nouns are removed, Muse Glimmer 30B provides the same answer for 96--99\% of MNLI, ANLI R1, and RTE items, and Qwen3.8-27B for 90\% of RTE items, compared to 51--64\% with clean input. Therefore, part of the decrease in accuracy is due to changes in the decision threshold rather than information loss.

Recentring the answer scores recovers up to 10 points (Qwen3.8-27B on RTE), and threshold-independent AUC shows the same results. The OLMo 3 models are less affected, and the three measures show agreement within approximately 2 points in ANLI R1 and RTE. For three of the four models, the ranking of tasks is the same under all three measures, with the least loss on MNLI and the greatest on ANLI R1 and RTE; the exception is Muse Glimmer 30B, whose raw MNLI accuracy is already low on clean input because of its answer bias.

\subsection{Hosted Models}

GPT-4o, scored by centred accuracy from its answer-token log-probabilities, behaves like the open-weight LLMs (Table~\ref{tab:main}), answer bias included: without nouns, the share of MNLI items receiving the most frequent answer rises from 38\% to 65\%, and recentring reduces the accuracy drop from 28.2 to 20.9 points.
DeepSeek-V4-Pro shows the strongest answer bias of all systems. Without nouns, it gives the same answer for 87\% of MNLI items and 93\% of RTE items, compared to 52\% and 59\% with clean input.
Judged by raw accuracy alone, it would keep only 22\% of its above-chance MNLI performance, which is misleading: centred accuracy gives 49\%.

\paragraph{A reasoning model.}
Grok 4.7, which reasons before answering, is the only system that does not follow the pattern.
It does better on clean ANLI than on clean MNLI (77.9\% vs.\ 71.8\%), and without nouns it keeps about as much of its performance on ANLI as on MNLI (38\% vs.\ 43\%), whereas every other system keeps at least twice as much on MNLI.
Since Grok 4.7 returns no answer scores, these are raw accuracies of a single model on the first 400 items of each task, and they should be read with caution.

\section{Memorisation or Shortcuts?}
\label{sec:contamination}

Why do all systems keep so much more of their performance on MNLI than on ANLI?
Given that MNLI is old and widely distributed, and thus a likely candidate for pre-training leakage, memorisation is a natural initial hypothesis.
We first test this with three contamination indicators and then examine the alternative, that MNLI entries can be resolved through cues that survive the perturbations.

\subsection{Contamination Indicators}

\paragraph{Guessing masked words.}
Following \citet{deng-etal-2024-investigating}, we masked one content word in each entry and asked Qwen3.8-27B to recover it.
In the initial version of this experiment, we allowed the masked word to appear in the other paired sentence, resulting in an exact match rate of 43--70\% in the sentence-pair tasks (Table~\ref{tab:contamination}).
However, most of these matches are obtained simply by copying across the pair.
If masking is restricted to words that do not appear elsewhere within the item, the exact match rate drops to 23--35\%, whereas there is no change for the single-sentence tasks CoLA and SST-2.
In these stricter settings, specific benchmarks do not stand out, and ANLI (released in 2020) is recovered with nearly the same frequency as MNLI.

\paragraph{Likelihood of rare tokens.}
Min-K\%++ \citep{zhang-etal-2025-mink} scores each item using the normalised log probability assigned to the least likely tokens in each item.
For OLMo 3 32B, the highest values appear in ANLI R1 and R2 ($-1.07$ and $-1.03$), while for the remaining tasks, they range from $-1.49$ to $-2.25$. In contrast, Qwen3.8-27B does not show a similar pattern.
Since ANLI R1 and R2 are based on Wikipedia passages, these high scores appear to reflect familiarity with the source text rather than exposure to the benchmark itself.

\paragraph{Exact $n$-gram counts.}
Since OLMo's pre-training corpus is publicly available, we can also check overlap directly.
Using the infini-gram API \citep{liu-etal-2024-infinigram}, we counted two evenly spaced 13-grams of each field (and the entire field, if up to 50 words long) in the OLMo~2 pre-training mix.\footnote{Index \texttt{v4\_olmo-mix-1124\_llama}; the OLMo~3 index was not yet available, so the counts are a lower bound for OLMo~3.}
A field counts as found if one of its 13-grams appears, or if the entire field appears and has at least 13 words, since short sentences such as \emph{The man is asleep.} appear in any large corpus.
The counts support the Min-K\%++ reading.
Premises were found in 72\% of ANLI R1 entries and 71\% of ANLI R2 entries, whereas in MNLI, this figure was only 21\%. Conversely, long hypotheses rarely appeared (1--4\% for both MNLI and ANLI).
Overall, the OLMo pre-training data contain a large part of the source texts of ANLI, but the benchmark entries themselves appear only rarely.

\subsection{Seen and Unseen Items}

If memorisation helps the model process perturbed inputs, items that appeared as text during pre-training should have less performance degradation than unseen items.
For each open-weight LLM, we compared the degradation of items with found premises and of the remaining items on MNLI and the three ANLI rounds, using centred accuracy and five perturbations.
Among all \premGapsAll{} comparisons, \premGapsExcl{} have 95\% confidence intervals excluding zero, with differences in both directions and an overall average of \premGapMean{} points.
We therefore find no evidence that exposure to the source text increases robustness.
For found hypotheses, the number of items (10--18 per task) is too small to support reliable comparisons of the same type.

\subsection{Hypothesis-Only Shortcuts}

\begin{figure}[t]
\centering
\includegraphics[width=\columnwidth]{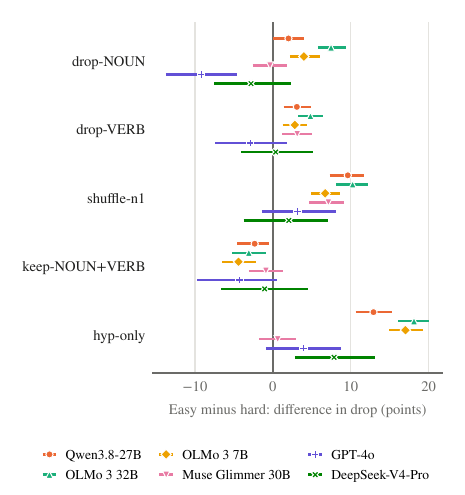}
\caption{MNLI degradation for items correctly classified by the hypothesis-only classifier (easy items) minus the degradation for the remaining items (hard items), expressed as centred accuracy points with 95\% bootstrap intervals. Positive values indicate less degradation for easy items. GPT-4o and DeepSeek-V4-Pro were evaluated on a sample of 2,000 items.}
\label{fig:shortcuts}
\end{figure}

Following \citet{gururangan-etal-2018-annotation}, we train a hypothesis-only classifier (logistic regression on word unigrams and bigrams) on the training hypotheses of each dataset, and split the evaluation items into correctly predicted items (easy items) and the remaining items (hard items).
The classifier reaches 54.3\% on MNLI, in line with \citet{gururangan-etal-2018-annotation}, 42.0--43.1\% on the ANLI rounds and 47.7\% (below chance) on RTE.
If hypothesis cues are the factor maintaining accuracy for perturbed inputs, then easy items should have a lower drop than hard items, and such a difference should depend on whether the perturbation preserves those cues.

Both predictions hold on MNLI for the three models with the most robust MNLI performance, Qwen3.8-27B and the two OLMo 3 models (Figure~\ref{fig:shortcuts}).
For OLMo 3 32B and 7B, easy items lose \scBigNoun{} and \scSmallNoun{} points less than hard items when nouns are removed, \scBigShuf{} and \scSmallShuf{} points less when words are shuffled, and \scBigHyp{} and \scSmallHyp{} points less when the premise is removed altogether.
Qwen3.8-27B shows the same pattern with a smaller gap without nouns (2.0 points), and all of these intervals exclude zero.
The one perturbation that reverses the pattern is \corr{keep-NOUN+VERB}: here easy items lose \emph{more}, by 2.3--4.4 points for the three models.
This is also the only perturbation that removes function words, including negation words such as \emph{not}, \emph{no} and \emph{never}, which are the strongest hypothesis-only cues in MNLI \citep{gururangan-etal-2018-annotation}.
Easy items are also easier on clean input (93.6\% vs.\ 87.4\% for OLMo 3 32B), which leaves them more room to drop, so the comparison is, if anything, conservative.

However, this pattern does not appear in all models.
For Muse Glimmer 30B, this phenomenon occurs only under the \corr{drop-VERB} and \corr{shuffle-n1} conditions. For GPT-4o, easy items show greater score loss (9.2 points) when nouns are absent, while no reliable difference appears under other conditions.
DeepSeek-V4-Pro shows a reliable difference (7.9 points) only when the premise is removed, and no difference appears in Mistral Large 3, which is scored based on raw answers.
The same applies to Grok 4.7 (all four MNLI intervals include 0). In other words, this effect is most clearly evident in open-weight models that we score ourselves, but not in hosted models.
In ANLI, results are mixed across perturbations, but when the premise is removed, most models show less R2 and R3 score loss for easier items (a difference of 8--14 points). This suggests that the weak cues remaining in ANLI are used in the same way.

\subsection{Fresh Items}
\label{sec:fresh}

\begin{table}[t]
\centering
\small
\setlength{\tabcolsep}{2.5pt}
\resizebox{\columnwidth}{!}{%
\begin{tabular}{lccccccc}
\toprule
& \multicolumn{2}{c}{Clean} & \multicolumn{2}{c}{$R$ no nouns} & \multicolumn{2}{c}{$R$ nouns+verbs} & Easy$-$ \\
\cmidrule(lr){2-3}\cmidrule(lr){4-5}\cmidrule(lr){6-7}
Model & MNLI & free & MNLI & free & MNLI & free & hard \\
\midrule
Qwen3.8-27B & 100.0 & 97.3 & 72 & 47 & 49 & 71 & +16.0$^*$ \\
OLMo 3 32B & 99.4 & 98.0 & 75 & 57 & 50 & 62 & +10.9 \\
OLMo 3 7B & 98.2 & 94.6 & 73 & 54 & 59 & 74 & +17.8$^*$ \\
Muse Glimmer 30B & 98.8 & 98.0 & 54 & 45 & 50 & 68 & +18.5$^*$ \\
\midrule
GPT-4o & 98.2 & 94.6 & 60 & 51 & 56 & 69 & +17.1$^*$ \\
DeepSeek-V4-Pro & 99.4 & 96.6 & 53 & 49 & 46 & 70 & +17.6$^*$ \\
Mistral Large 3$^\dagger$ & 98.2 & 93.2 & 52 & 37 & 42 & 68 & +28.1$^*$ \\
Grok 4.7 (reasoning)$^{\dagger\S}$ & 98.2 & 93.9 & 50 & 42 & -- & -- & +22.7$^*$ \\
 
\bottomrule
\end{tabular}}
\caption{New items written in MNLI style or without cues (168 and 148 items): clean accuracy (\%, centred where answer scores exist), retention rate of performance above chance ($R$, \%) under \corr{drop-NOUN} and \corr{keep-NOUN+VERB} conditions, and difference between easy and hard items in \corr{drop-NOUN} drop for MNLI-style items (points, with $^*$ for 95\% confidence intervals excluding 0). Symbols are the same as Table~\ref{tab:main}.}
\label{tab:fresh}
\end{table}

The new items enable the most direct test: no model has seen them, but the MNLI-style items have the same type of cues as MNLI.
If memorisation causes the robustness on MNLI, it should disappear on these items; if shortcuts cause it, it should remain on MNLI-style items and shrink on cue-free ones.

Table~\ref{tab:fresh} shows the second pattern in all models.
In the absence of nouns, all eight LLMs maintain higher performance in MNLI-style items (50--75\% vs.\ 37--57\%) than in cue-free items, which is nearly the same level as performance in MNLI itself.
When only nouns and verbs remain, the order is reversed in all models tested with this perturbation (42--59\% vs.\ 62--74\%), as on MNLI.
Within MNLI-style items, items correctly labelled by the MNLI hypothesis-only classifier show 11--28 points less loss than the remaining items when nouns are absent, and for 7 of the 8 models the 95\% interval excludes zero.
The same effect is observed here in the reasoning model and in hosted models, even though such effects were not present on MNLI.
The new items are easier than MNLI (93--100\% clean accuracy), and since most were labelled by a single annotator, absolute figures should be interpreted with caution. However, the difference between the two styles is consistently observed across all models.

\subsection{A Benchmark Without Hypothesis Cues}
\label{sec:bioasq}

Robustness to perturbations does not necessarily come from shortcuts: models may also answer from what they already know.
To see robustness without hypothesis cues, we perturbed the evidence passages of the 1,519 yes/no questions in BioASQ \citep{tsatsaronis-etal-2015-overview} and kept the questions, which act as hypotheses, intact.
A classifier using only the questions did not outperform the majority baseline on the three most recent test sets, and we found no sign of memorisation (Appendix~\ref{app:bioasq}).
Nevertheless, the models retained 64--89\% of their above-chance AUC without any evidence, compared to 26--51\% on MNLI using only hypotheses (Table~\ref{tab:bioasq} in Appendix~\ref{app:bioasq}).
Unlike MNLI, keeping only nouns and verbs is almost no more costly than removing nouns (83--96\% retention compared to 38--47\% retention in MNLI).
BioASQ is therefore robust for other reasons, most likely biomedical knowledge, and the reversal under \corr{keep-NOUN+VERB} in MNLI appears to be specific to data whose cues lie in function words.

\section{Discussion}

Taken together, the results point away from memorisation and towards shortcuts as the reason why familiar NLI benchmarks survive perturbation.
Four observations support this reading.
First, the contrast between MNLI and ANLI is shared by every system that answers directly, including RoBERTa fine-tuned on MNLI, whose robustness comes from what it learned from MNLI's training data rather than from having seen the evaluation items.
Second, none of the contamination indicators is related to robustness: items whose text is found in OLMo's pre-training data lose as much as items whose text is not.
Third, for the open LLMs most robust on MNLI, items that can be solved from the hypothesis alone are more robust, and lose this advantage exactly when the perturbation removes the function words that carry the cues.
Fourth, and most directly, every LLM shows the same signature on fresh items that it cannot have seen, as long as they are written in MNLI's style.

The hosted models and the reasoning model show no hypothesis-only effect on MNLI itself, although they do on the fresh MNLI-style items, so they can exploit the same cues.
Why the effect is absent on MNLI remains open: their training data are unknown, so memorisation of MNLI cannot be ruled out for them, and they may rely on cues our classifier does not capture, such as lexical overlap \citep{mccoy-etal-2019-right}.

On MNLI, a large part of the benchmark can be solved from surface properties of the hypothesis that the perturbations leave intact; ANLI was built to remove such properties, and on ANLI all systems fall close to chance once the nouns are gone.
In each case the perturbation experiments measure the benchmark as much as the model.

Do models, then, need to understand language to do well on these benchmarks?
Our results suggest that the answer depends on the benchmark more than on the model.
On MNLI, understanding the sentence pair is sufficient for a high score but not necessary: much of the benchmark can still be solved once the words a human reader would need are gone \citep{talman-etal-2021-nli}.
In this sense MNLI measures, in part, how its annotators wrote hypotheses rather than whether one sentence follows from another.
This does not mean that the models fail to understand the pairs.
On the cue-free fresh items every LLM reaches 93--98\% on clean input, so the models can draw the inference when no cue is available; a benchmark on which a cue-based and an inference-based solution score alike simply cannot tell the two apart.
BioASQ adds a further distinction, between understanding the evidence and already knowing the answer.
We would therefore argue that a benchmark score neither confirms nor refutes understanding, which is better seen as a matter of degree, and that perturbation experiments are most useful for finding out what a given benchmark can and cannot distinguish.

\section{Conclusion}

In this paper we have revisited the effect of input perturbations on NLI and other language understanding tasks with LLMs and fine-tuned encoders.
Every system that answers directly keeps 36--72\% of its above-chance MNLI performance without nouns but only 14--29\% on ANLI, whatever its size.
Contamination indicators show no link between exposure and robustness, whereas items solvable from the hypothesis are more robust.
On 368 fresh items that no model can have seen, all eight LLMs keep more of their performance on items written in MNLI's style than on cue-free items, and lose this advantage when function words are removed.
We therefore argue that robustness to perturbation on familiar NLI benchmarks reflects shortcuts in the datasets rather than memorisation.
We recommend that robustness results on NLI benchmarks be reported together with a hypothesis-only analysis and, where possible, on items written after the models' training data were collected.

Open questions remain: a larger, multiply annotated fresh set would sharpen the comparison; other cues, such as lexical overlap \citep{mccoy-etal-2019-right}, may explain why the hosted models show no hypothesis-only effect on MNLI; and it remains to be seen whether reasoning models in general escape the MNLI--ANLI contrast.

\section*{Limitations}

Our contamination checks are partial.
The $n$-gram counts cover the OLMo~2 pre-training mix rather than that of OLMo~3, sample only two 13-grams per field, and detect only verbatim overlap; paraphrased or reformatted copies of a benchmark, for example in code repositories, would be missed.
For the other models the training data are not available, and the indirect indicators we use are known to be noisy.
The absence of a link between exposure and robustness is therefore evidence against memorisation as the main explanation, not proof that the models have not seen the benchmarks; the fresh items address this for the comparison that matters most, but not for MNLI itself.

The shortcut split depends on one simple classifier; items it labels incorrectly may still contain cues that a stronger classifier would find, so the split underestimates the number of shortcut-solvable items, and a stronger hypothesis-only model, such as a fine-tuned encoder, would give a sharper one.
The hosted models are analysed on the 2,000-item MNLI sample, which gives the comparison less power than for the open models.

The LLMs are scored with a single prompt template per task and, for the open models, by answer log-probability.
Other templates or free-form generation could give different absolute numbers, although the comparisons we draw are within a model and template; a second template would show how far the absolute numbers depend on it.
The hosted models are evaluated on the core set of perturbations and, for MNLI and QQP, on a 2,000-item sample, and their providers may change the models behind an endpoint over time.
All experiments are on English data.
The BioASQ comparison covers only the yes/no questions and six models, and is scored by AUC rather than centred accuracy.
The fresh items were written by two LLMs and validated mostly by a single volunteer annotator each, none a native English speaker; their hypotheses differ between the two styles in more than cues (for example, cue-free hypotheses are longer), and a larger set with several labels per item would make the comparison more reliable.

\section*{Ethics Statement}
We use GLUE \citep{wang-etal-2018-glue} under the licences of its component datasets, ANLI under CC BY-NC 4.0, and Wikipedia text under CC BY-SA 4.0; open models are used under their published licences and hosted models under their providers' terms. We report results only in aggregate.
The fresh items were labelled by unpaid volunteers who could stop at any time; their age and language background were collected only to describe the rater pool and are reported only in aggregate.
Hosted models were accessed through their providers' standard interfaces; a small number of items (at most 0.5\% per perturbation for MNLI and ANLI, 1\% for RTE) were blocked by a provider's content filter and are counted as errors.

\ifarxiv\section*{Acknowledgements}
We thank the volunteers who labelled the fresh items.
We acknowledge CSC -- IT Center for Science, Finland, for computational resources on the Roihu supercomputer.
The second author gratefully acknowledges support from the Human Diversity consortium, under the Profi7 program (grant no.~352727) and the Centre of Excellence (grant no.~374221) by the Research Council of Finland.
\fi

\bibliography{references}

@inproceedings{talman-etal-2021-nli,
    title = "{NLI} Data Sanity Check: Assessing the Effect of Data Corruption on Model Performance",
    author = {Talman, Aarne  and
      Apidianaki, Marianna  and
      Chatzikyriakidis, Stergios  and
      Tiedemann, J{\"o}rg},
    editor = "Dobnik, Simon  and
      {\O}vrelid, Lilja",
    booktitle = "Proceedings of the 23rd {N}ordic Conference on Computational Linguistics ({N}o{D}a{L}i{D}a)",
    month = may # " 31--2 " # jun,
    year = "2021",
    address = "Reykjavik, Iceland (Online)",
    publisher = {Link{\"o}ping University Electronic Press, Sweden},
    url = "https://aclanthology.org/2021.nodalida-main.28/",
    pages = "276--287"
}

@inproceedings{talman-etal-2022-data,
    title = "How Does Data Corruption Affect Natural Language Understanding Models? A Study on {GLUE} datasets",
    author = {Talman, Aarne  and
      Apidianaki, Marianna  and
      Chatzikyriakidis, Stergios  and
      Tiedemann, J{\"o}rg},
    editor = "Nastase, Vivi  and
      Pavlick, Ellie  and
      Pilehvar, Mohammad Taher  and
      Camacho-Collados, Jose  and
      Raganato, Alessandro",
    booktitle = "Proceedings of the 11th Joint Conference on Lexical and Computational Semantics",
    month = jul,
    year = "2022",
    address = "Seattle, Washington",
    publisher = "Association for Computational Linguistics",
    url = "https://aclanthology.org/2022.starsem-1.20/",
    doi = "10.18653/v1/2022.starsem-1.20",
    pages = "226--233"
}

@inproceedings{gururangan-etal-2018-annotation,
    title = "Annotation Artifacts in Natural Language Inference Data",
    author = "Gururangan, Suchin  and
      Swayamdipta, Swabha  and
      Levy, Omer  and
      Schwartz, Roy  and
      Bowman, Samuel R.  and
      Smith, Noah A.",
    editor = "Walker, Marilyn  and
      Ji, Heng  and
      Stent, Amanda",
    booktitle = "Proceedings of the 2018 Conference of the {N}orth {A}merican Chapter of the {A}ssociation for {C}omputational {L}inguistics: Human Language Technologies, Volume 2 (Short Papers)",
    month = jun,
    year = "2018",
    address = "New Orleans, Louisiana",
    publisher = "Association for Computational Linguistics",
    url = "https://aclanthology.org/N18-2017/",
    doi = "10.18653/v1/N18-2017",
    pages = "107--112"
}

@inproceedings{poliak-etal-2018-hypothesis,
    title = "Hypothesis Only Baselines in Natural Language Inference",
    author = "Poliak, Adam  and
      Naradowsky, Jason  and
      Haldar, Aparajita  and
      Rudinger, Rachel  and
      Van Durme, Benjamin",
    editor = "Nissim, Malvina  and
      Berant, Jonathan  and
      Lenci, Alessandro",
    booktitle = "Proceedings of the Seventh Joint Conference on Lexical and Computational Semantics",
    month = jun,
    year = "2018",
    address = "New Orleans, Louisiana",
    publisher = "Association for Computational Linguistics",
    url = "https://aclanthology.org/S18-2023/",
    doi = "10.18653/v1/S18-2023",
    pages = "180--191"
}

@inproceedings{mccoy-etal-2019-right,
    title = "Right for the Wrong Reasons: Diagnosing Syntactic Heuristics in Natural Language Inference",
    author = "McCoy, R. Thomas  and
      Pavlick, Ellie  and
      Linzen, Tal",
    editor = "Korhonen, Anna  and
      Traum, David  and
      M{\`a}rquez, Llu{\'i}s",
    booktitle = "Proceedings of the 57th Annual Meeting of the {A}ssociation for {C}omputational {L}inguistics",
    month = jul,
    year = "2019",
    address = "Florence, Italy",
    publisher = "Association for Computational Linguistics",
    url = "https://aclanthology.org/P19-1334/",
    doi = "10.18653/v1/P19-1334",
    pages = "3428--3448"
}

@inproceedings{williams-etal-2018-broad,
    title = "A Broad-Coverage Challenge Corpus for Sentence Understanding through Inference",
    author = "Williams, Adina  and
      Nangia, Nikita  and
      Bowman, Samuel R.",
    editor = "Walker, Marilyn  and
      Ji, Heng  and
      Stent, Amanda",
    booktitle = "Proceedings of the 2018 Conference of the {N}orth {A}merican Chapter of the {A}ssociation for {C}omputational {L}inguistics: Human Language Technologies, Volume 1 (Long Papers)",
    month = jun,
    year = "2018",
    address = "New Orleans, Louisiana",
    publisher = "Association for Computational Linguistics",
    url = "https://aclanthology.org/N18-1101/",
    doi = "10.18653/v1/N18-1101",
    pages = "1112--1122"
}

@inproceedings{nie-etal-2020-adversarial,
    title = "Adversarial {NLI}: A New Benchmark for Natural Language Understanding",
    author = "Nie, Yixin  and
      Williams, Adina  and
      Dinan, Emily  and
      Bansal, Mohit  and
      Weston, Jason  and
      Kiela, Douwe",
    editor = "Jurafsky, Dan  and
      Chai, Joyce  and
      Schluter, Natalie  and
      Tetreault, Joel",
    booktitle = "Proceedings of the 58th Annual Meeting of the {A}ssociation for {C}omputational {L}inguistics",
    month = jul,
    year = "2020",
    address = "Online",
    publisher = "Association for Computational Linguistics",
    url = "https://aclanthology.org/2020.acl-main.441/",
    doi = "10.18653/v1/2020.acl-main.441",
    pages = "4885--4901"
}

@inproceedings{wang-etal-2018-glue,
    title = "{GLUE}: A Multi-Task Benchmark and Analysis Platform for Natural Language Understanding",
    author = "Wang, Alex  and
      Singh, Amanpreet  and
      Michael, Julian  and
      Hill, Felix  and
      Levy, Omer  and
      Bowman, Samuel R.",
    editor = "Linzen, Tal  and
      Chrupa{\l}a, Grzegorz  and
      Alishahi, Afra",
    booktitle = "Proceedings of the 2018 {EMNLP} Workshop {B}lackbox{NLP}: Analyzing and Interpreting Neural Networks for {NLP}",
    month = nov,
    year = "2018",
    address = "Brussels, Belgium",
    publisher = "Association for Computational Linguistics",
    url = "https://aclanthology.org/W18-5446/",
    doi = "10.18653/v1/W18-5446",
    pages = "353--355"
}

@inproceedings{sinha-etal-2021-unnatural,
    title = "{UnNatural} {L}anguage {I}nference",
    author = "Sinha, Koustuv  and
      Parthasarathi, Prasanna  and
      Pineau, Joelle  and
      Williams, Adina",
    editor = "Zong, Chengqing  and
      Xia, Fei  and
      Li, Wenjie  and
      Navigli, Roberto",
    booktitle = "Proceedings of the 59th Annual Meeting of the {A}ssociation for {C}omputational {L}inguistics and the 11th International Joint Conference on Natural Language Processing (Volume 1: Long Papers)",
    month = aug,
    year = "2021",
    address = "Online",
    publisher = "Association for Computational Linguistics",
    url = "https://aclanthology.org/2021.acl-long.569/",
    doi = "10.18653/v1/2021.acl-long.569",
    pages = "7329--7346"
}

@inproceedings{sinha-etal-2021-masked,
    title = "Masked Language Modeling and the Distributional Hypothesis: Order Word Matters Pre-training for Little",
    author = "Sinha, Koustuv  and
      Jia, Robin  and
      Hupkes, Dieuwke  and
      Pineau, Joelle  and
      Williams, Adina  and
      Kiela, Douwe",
    editor = "Moens, Marie-Francine  and
      Huang, Xuanjing  and
      Specia, Lucia  and
      Yih, Scott Wen-tau",
    booktitle = "Proceedings of the 2021 Conference on Empirical Methods in Natural Language Processing",
    month = nov,
    year = "2021",
    address = "Online and Punta Cana, Dominican Republic",
    publisher = "Association for Computational Linguistics",
    url = "https://aclanthology.org/2021.emnlp-main.230/",
    doi = "10.18653/v1/2021.emnlp-main.230",
    pages = "2888--2913"
}

@inproceedings{pham-etal-2021-order,
    title = "Out of Order: How important is the sequential order of words in a sentence in Natural Language Understanding tasks?",
    author = "Pham, Thang  and
      Bui, Trung  and
      Mai, Long  and
      Nguyen, Anh",
    editor = "Zong, Chengqing  and
      Xia, Fei  and
      Li, Wenjie  and
      Navigli, Roberto",
    booktitle = "Findings of the {A}ssociation for {C}omputational {L}inguistics: {ACL}-{IJCNLP} 2021",
    month = aug,
    year = "2021",
    address = "Online",
    publisher = "Association for Computational Linguistics",
    url = "https://aclanthology.org/2021.findings-acl.98/",
    doi = "10.18653/v1/2021.findings-acl.98",
    pages = "1145--1160"
}

@inproceedings{gupta-etal-2021-bert,
  title = {{BERT} \& Family Eat Word Salad: Experiments with Text Understanding},
  author = {Gupta, Ashim and Kvernadze, Giorgi and Srikumar, Vivek},
  booktitle = {Proceedings of the AAAI Conference on Artificial Intelligence},
  volume = {35},
  number = {14},
  pages = {12946--12954},
  year = {2021},
  doi = {10.1609/aaai.v35i14.17531}
}

@inproceedings{zhao-etal-2021-calibrate,
  title = {Calibrate Before Use: Improving Few-Shot Performance of Language Models},
  author = {Zhao, Zihao and Wallace, Eric and Feng, Shi and Klein, Dan and Singh, Sameer},
  booktitle = {Proceedings of the 38th International Conference on Machine Learning},
  year = {2021}
}

@inproceedings{holtzman-etal-2021-surface,
    title = "Surface Form Competition: Why the Highest Probability Answer Isn{'}t Always Right",
    author = "Holtzman, Ari  and
      West, Peter  and
      Shwartz, Vered  and
      Choi, Yejin  and
      Zettlemoyer, Luke",
    editor = "Moens, Marie-Francine  and
      Huang, Xuanjing  and
      Specia, Lucia  and
      Yih, Scott Wen-tau",
    booktitle = "Proceedings of the 2021 Conference on Empirical Methods in Natural Language Processing",
    month = nov,
    year = "2021",
    address = "Online and Punta Cana, Dominican Republic",
    publisher = "Association for Computational Linguistics",
    url = "https://aclanthology.org/2021.emnlp-main.564/",
    doi = "10.18653/v1/2021.emnlp-main.564",
    pages = "7038--7051"
}

@inproceedings{shi-etal-2024-detecting,
  title = {Detecting Pretraining Data from Large Language Models},
  author = {Shi, Weijia and Ajith, Anirudh and Xia, Mengzhou and Huang, Yangsibo and Liu, Daogao and Blevins, Terra and Chen, Danqi and Zettlemoyer, Luke},
  booktitle = {The Twelfth International Conference on Learning Representations},
  year = {2024}
}

@inproceedings{zhang-etal-2025-mink,
  title = {Min-K\%++: Improved Baseline for Detecting Pre-Training Data from Large Language Models},
  author = {Zhang, Jingyang and Sun, Jingwei and Yeats, Eric and Ouyang, Yang and Kuo, Martin and Zhang, Jianyi and Yang, Hao Frank and Li, Hai},
  booktitle = {The Thirteenth International Conference on Learning Representations},
  year = {2025}
}

@inproceedings{deng-etal-2024-investigating,
    title = "Investigating Data Contamination in Modern Benchmarks for Large Language Models",
    author = "Deng, Chunyuan  and
      Zhao, Yilun  and
      Tang, Xiangru  and
      Gerstein, Mark  and
      Cohan, Arman",
    editor = "Duh, Kevin  and
      Gomez, Helena  and
      Bethard, Steven",
    booktitle = "Proceedings of the 2024 Conference of the {N}orth {A}merican Chapter of the {A}ssociation for {C}omputational {L}inguistics: Human Language Technologies (Volume 1: Long Papers)",
    month = jun,
    year = "2024",
    address = "Mexico City, Mexico",
    publisher = "Association for Computational Linguistics",
    url = "https://aclanthology.org/2024.naacl-long.482/",
    doi = "10.18653/v1/2024.naacl-long.482",
    pages = "8706--8719"
}

@inproceedings{liu-etal-2024-infinigram,
  title = {Infini-gram: Scaling Unbounded n-gram Language Models to a Trillion Tokens},
  author = {Liu, Jiacheng and Min, Sewon and Zettlemoyer, Luke and Choi, Yejin and Hajishirzi, Hannaneh},
  booktitle = {First Conference on Language Modeling},
  year = {2024}
}

@inproceedings{sainz-etal-2023-nlp,
    title = "{NLP} Evaluation in trouble: On the Need to Measure {LLM} Data Contamination for each Benchmark",
    author = "Sainz, Oscar  and
      Campos, Jon  and
      Garc{\'i}a-Ferrero, Iker  and
      Etxaniz, Julen  and
      de Lacalle, Oier Lopez  and
      Agirre, Eneko",
    editor = "Bouamor, Houda  and
      Pino, Juan  and
      Bali, Kalika",
    booktitle = "Findings of the {A}ssociation for {C}omputational {L}inguistics: {EMNLP} 2023",
    month = dec,
    year = "2023",
    address = "Singapore",
    publisher = "Association for Computational Linguistics",
    url = "https://aclanthology.org/2023.findings-emnlp.722/",
    doi = "10.18653/v1/2023.findings-emnlp.722",
    pages = "10776--10787"
}

@inproceedings{golchin-surdeanu-2024-time,
  title = {Time Travel in {LLM}s: Tracing Data Contamination in Large Language Models},
  author = {Golchin, Shahriar and Surdeanu, Mihai},
  booktitle = {The Twelfth International Conference on Learning Representations},
  year = {2024}
}

@article{liu-etal-2019-roberta,
  title = {{RoBERTa}: A Robustly Optimized {BERT} Pretraining Approach},
  author = {Liu, Yinhan and Ott, Myle and Goyal, Naman and Du, Jingfei and Joshi, Mandar and Chen, Danqi and Levy, Omer and Lewis, Mike and Zettlemoyer, Luke and Stoyanov, Veselin},
  journal = {arXiv preprint arXiv:1907.11692},
  year = {2019}
}

@inproceedings{devlin-etal-2019-bert,
    title = "{BERT}: Pre-training of Deep Bidirectional Transformers for Language Understanding",
    author = "Devlin, Jacob  and
      Chang, Ming-Wei  and
      Lee, Kenton  and
      Toutanova, Kristina",
    editor = "Burstein, Jill  and
      Doran, Christy  and
      Solorio, Thamar",
    booktitle = "Proceedings of the 2019 Conference of the {N}orth {A}merican Chapter of the {A}ssociation for {C}omputational {L}inguistics: Human Language Technologies, Volume 1 (Long and Short Papers)",
    month = jun,
    year = "2019",
    address = "Minneapolis, Minnesota",
    publisher = "Association for Computational Linguistics",
    url = "https://aclanthology.org/N19-1423/",
    doi = "10.18653/v1/N19-1423",
    pages = "4171--4186"
}

@inproceedings{conneau-etal-2020-unsupervised,
    title = "Unsupervised Cross-lingual Representation Learning at Scale",
    author = "Conneau, Alexis  and
      Khandelwal, Kartikay  and
      Goyal, Naman  and
      Chaudhary, Vishrav  and
      Wenzek, Guillaume  and
      Guzm{\'a}n, Francisco  and
      Grave, Edouard  and
      Ott, Myle  and
      Zettlemoyer, Luke  and
      Stoyanov, Veselin",
    editor = "Jurafsky, Dan  and
      Chai, Joyce  and
      Schluter, Natalie  and
      Tetreault, Joel",
    booktitle = "Proceedings of the 58th Annual Meeting of the {A}ssociation for {C}omputational {L}inguistics",
    month = jul,
    year = "2020",
    address = "Online",
    publisher = "Association for Computational Linguistics",
    url = "https://aclanthology.org/2020.acl-main.747/",
    doi = "10.18653/v1/2020.acl-main.747",
    pages = "8440--8451"
}

@inproceedings{warner-etal-2025-smarter,
    title = "Smarter, Better, Faster, Longer: A Modern Bidirectional Encoder for Fast, Memory Efficient, and Long Context Finetuning and Inference",
    author = {Warner, Benjamin  and
      Chaffin, Antoine  and
      Clavi{\'e}, Benjamin  and
      Weller, Orion  and
      Hallstr{\"o}m, Oskar  and
      Taghadouini, Said  and
      Gallagher, Alexis  and
      Biswas, Raja  and
      Ladhak, Faisal  and
      Aarsen, Tom  and
      Adams, Griffin Thomas  and
      Howard, Jeremy  and
      Poli, Iacopo},
    editor = "Che, Wanxiang  and
      Nabende, Joyce  and
      Shutova, Ekaterina  and
      Pilehvar, Mohammad Taher",
    booktitle = "Proceedings of the 63rd Annual Meeting of the {A}ssociation for {C}omputational {L}inguistics (Volume 1: Long Papers)",
    month = jul,
    year = "2025",
    address = "Vienna, Austria",
    publisher = "Association for Computational Linguistics",
    url = "https://aclanthology.org/2025.acl-long.127/",
    doi = "10.18653/v1/2025.acl-long.127",
    pages = "2526--2547",
    ISBN = "979-8-89176-251-0"
}

@inproceedings{petrov-etal-2012-universal,
    title = "A Universal Part-of-Speech Tagset",
    author = "Petrov, Slav  and
      Das, Dipanjan  and
      McDonald, Ryan",
    editor = "Calzolari, Nicoletta  and
      Choukri, Khalid  and
      Declerck, Thierry  and
      Do{\u{g}}an, Mehmet U{\u{g}}ur  and
      Maegaard, Bente  and
      Mariani, Joseph  and
      Moreno, Asuncion  and
      Odijk, Jan  and
      Piperidis, Stelios",
    booktitle = "Proceedings of the Eighth International Conference on Language Resources and Evaluation ({LREC}'12)",
    month = may,
    year = "2012",
    address = "Istanbul, Turkey",
    publisher = "European Language Resources Association (ELRA)",
    url = "https://aclanthology.org/L12-1115/",
    pages = "2089--2096"
}

@book{bird-etal-2009-nltk,
  title = {Natural Language Processing with Python},
  author = {Bird, Steven and Klein, Ewan and Loper, Edward},
  publisher = {O'Reilly Media},
  year = {2009}
}

@article{tsatsaronis-etal-2015-overview,
    title = "An overview of the {BIOASQ} large-scale biomedical semantic indexing and question answering competition",
    author = "Tsatsaronis, George and Balikas, Georgios and Malakasiotis, Prodromos and Partalas, Ioannis and Zschunke, Matthias and Alvers, Michael R. and Weissenborn, Dirk and Krithara, Anastasia and Petridis, Sergios and Polychronopoulos, Dimitris and Almirantis, Yannis and Pavlopoulos, John and Baskiotis, Nicolas and Gallinari, Patrick and Arti{\`e}res, Thierry and Ngonga Ngomo, Axel-Cyrille and Heino, Norman and Gaussier, Eric and Barrio-Alvers, Liliana and Schroeder, Michael and Androutsopoulos, Ion and Paliouras, Georgios",
    journal = "BMC Bioinformatics",
    volume = "16",
    number = "1",
    pages = "138",
    year = "2015",
    doi = "10.1186/s12859-015-0564-6"
}

\appendix
\FloatBarrier
\section{Reproducing the Earlier Results}
\label{app:anli}
We reproduce the \corr{drop-NOUN} experiments of \citet{talman-etal-2022-data} with RoBERTa-base on five GLUE tasks and those of \citet{talman-etal-2021-nli} with BERT-base on MNLI.
As in the earlier work, there are three conditions: \cond{Corrupt-Test} (trained on clean data, tested on perturbed data), \cond{Corrupt-Train} (trained on perturbed data, tested on clean data) and \cond{Corrupt-Train-and-Test} (perturbed in both).
Table~\ref{tab:reproduction} gives the published and the reproduced scores side by side.
On MNLI, all reproduced scores are within 0.6 points of the published ones, and the other GLUE tasks are within about 3 points, with two exceptions.
Our clean CoLA score is 5.2 points lower, and our \cond{Corrupt-Train} models reach 62.9 on RTE instead of 47.3.
The published RTE figure equals the share of entailment items in the development set (131 of 277), which suggests that the original run predicted a single class; our three seeds did not.
For ANLI, we evaluate the released RoBERTa-large model of \citet{nie-etal-2020-adversarial} on the clean test sets and on the perturbed test sets released by \citet{talman-etal-2021-nli}; all scores are within three points of the published ones (Table~\ref{tab:anli-reproduction}).

\begin{table*}[h]
\centering
\small
\begin{tabular}{llcccc}
\toprule
Model & Task & Clean & \cond{Corrupt-Test} & \cond{Corrupt-Train} & \cond{Corrupt-Train-and-Test} \\
\midrule
RoBERTa-base & MNLI & 87.9 / 87.8 & 72.8 / 72.8 & 85.6 / 85.8 & 77.5 / 77.7 \\
RoBERTa-base & RTE & 70.0 / 70.8 & 53.8 / 54.0 & 47.3 / 62.9 & 47.3 / 56.7 \\
RoBERTa-base & MRPC & 88.7 / 88.1 & 82.3 / 79.6 & 86.3 / 82.0 & 80.2 / 82.3 \\
RoBERTa-base & CoLA & 64.0 / 58.8 & 17.8 / 15.3 & 39.7 / 41.3 & 34.3 / 34.9 \\
RoBERTa-base & STS-B & 90.1 / 89.9 & 56.1 / 55.8 & 81.7 / 81.8 & 63.5 / 63.3 \\
BERT-base & MNLI & 83.7 / 83.9 & 69.8 / 70.2 & 80.7 / 81.3 & 73.4 / 73.3 \\
 
\bottomrule
\end{tabular}
\caption{Published / reproduced scores for \corr{drop-NOUN}. RoBERTa-base rows are compared with \citet{talman-etal-2022-data}, BERT-base with \citet{talman-etal-2021-nli}. Metrics as in GLUE: accuracy (MNLI, RTE), F1 (MRPC), Matthews correlation (CoLA) and Pearson correlation (STS-B), all $\times100$; ours are means over three seeds.}
\label{tab:reproduction}
\end{table*}

\begin{table}[h]
\centering
\small
\resizebox{\columnwidth}{!}{%
\begin{tabular}{lccc}
\toprule
 & R1 & R2 & R3 \\
\midrule
Clean & 73.8 / 73.5 & 48.9 / 49.3 & 44.4 / 45.5 \\
\textsc{Corrupt-Test} NOUN & 43.7 / 46.5 & 36.0 / 34.7 & 32.4 / 33.8 \\
\textsc{Corrupt-Test} VERB & 54.6 / 57.2 & 44.7 / 42.3 & 39.3 / 39.0 \\
 
\bottomrule
\end{tabular}}
\caption{ANLI test accuracy (\%) of the released RoBERTa-large model of \citet{nie-etal-2020-adversarial}, published / ours. Clean scores are compared with \citet{nie-etal-2020-adversarial}, perturbed ones with \citet{talman-etal-2021-nli}, using that paper's released files.}
\label{tab:anli-reproduction}
\end{table}

\FloatBarrier
\section{Prompts}
\label{app:prompts}
All tasks share one template per task; the NLI tasks (MNLI, ANLI, fresh items) use:
\begin{quote}\small\ttfamily
Premise: \{premise\}\\
Hypothesis: \{hypothesis\}\\
Does the premise entail the hypothesis, contradict it, or neither?\\
Answer with one word: entailment, neutral, or contradiction.
\end{quote}
RTE asks \emph{Does the premise entail the hypothesis?} with the answers \emph{yes} or \emph{no}; QNLI asks whether a sentence contains the answer to a question; MRPC and QQP whether two sentences (questions) mean (ask) the same thing; SST-2 for the sentiment (\emph{positive} or \emph{negative}); CoLA whether a sentence is grammatically acceptable; STS-B for a similarity score from 0 to 5.
Base models receive the text followed by \texttt{Answer:}; chat models and hosted models receive it as a user message.
In the five-shot setting, five training examples precede the item, each followed by its gold answer (as \texttt{Answer: <word>} for base models, as an assistant turn for chat models), drawn label-balanced with a fixed seed; for the ablated variant the same perturbation is applied to the examples.

\FloatBarrier
\section{Ablated In-Context Examples}
\label{app:fewshot}
For encoders, training on ablated data recovers most of the loss \citep{talman-etal-2021-nli,talman-etal-2022-data}: RoBERTa-base reaches 85.8 on clean MNLI after training without nouns, and 77.7 when tested without nouns as well (Table~\ref{tab:reproduction}).
Is there an in-context analogue for LLMs?
Table~\ref{tab:fewshot} suggests that there is not.
Five examples help OLMo 3 32B on ANLI R1 and QNLI, raising the retained share from 23\% to 31\% and from 44\% to 56\%.
Ablated examples, however, show no consistent advantage over clean ones: the two differ by at most six points in either direction, and on MNLI not at all (72\% with either kind for OLMo 3 32B, 67\% for OLMo 3 7B).
The model does not adapt to the ablation shown in the examples; the gain comes from the examples themselves.

\begin{table*}[h]
\centering
\small
\begin{tabular}{llcccccc}
\toprule
Model & Examples & MNLI & RTE & ANLI R1 & ANLI R2 & ANLI R3 & QNLI \\
\midrule
OLMo 3 7B & 0-shot & 88.3 (68) & 89.2 (40) & 63.0 (25) & 50.1 (3) & 47.5 (13) & 82.0 (56) \\
 & 5-shot, clean & 88.4 (67) & 88.5 (35) & 66.2 (28) & 50.9 (18) & 50.2 (18) & 87.0 (56) \\
 & 5-shot, drop-NOUN & 87.5 (67) & 88.5 (33) & 63.6 (25) & 49.4 (12) & 49.6 (20) & 86.5 (57) \\
\midrule
OLMo 3 32B & 0-shot & 90.8 (70) & 91.3 (30) & 74.0 (23) & 61.1 (19) & 56.8 (17) & 88.5 (44) \\
 & 5-shot, clean & 89.0 (72) & 92.1 (33) & 75.0 (31) & 64.4 (19) & 60.8 (17) & 91.0 (56) \\
 & 5-shot, drop-NOUN & 89.2 (72) & 91.7 (34) & 75.1 (31) & 62.8 (22) & 60.8 (18) & 90.8 (59) \\
 
\bottomrule
\end{tabular}
\caption{Clean centred accuracy (\%) and, in brackets, the share of above-chance performance retained under \corr{drop-NOUN} ($R$, \%), with no examples, five clean examples or five examples with nouns removed. MNLI five-shot on the 2,000-item sample.}
\label{tab:fewshot}
\end{table*}

\FloatBarrier
\section{Full Results}
\label{app:full}
The tables in this section give the results behind Section~\ref{sec:results}: the effect of the answer bias (Table~\ref{tab:answer-bias}), the share retained under every perturbation for the LLMs (Table~\ref{tab:full-llm}) and the fine-tuned encoders (Table~\ref{tab:encoders}).

\begin{table}[h]
\centering
\small
\setlength{\tabcolsep}{4pt}
\resizebox{\columnwidth}{!}{%
\begin{tabular}{llcccc}
\toprule
Task & Model & $\Delta$Acc & $\Delta$Cen & $\Delta$AUC & Top answer \\
\midrule
MNLI & Qwen3.8-27B & -25.9 & -20.4 & -13.4 & 35 $\rightarrow$ 57 \\
 & OLMo 3 32B & -19.5 & -17.5 & -10.1 & 38 $\rightarrow$ 44 \\
 & OLMo 3 7B & -19.8 & -17.4 & -10.4 & 35 $\rightarrow$ 47 \\
 & Muse Glimmer 30B & -31.9 & -25.7 & -21.3 & 63 $\rightarrow$ 96 \\
\midrule
ANLI R1 & Qwen3.8-27B & -32.3 & -32.8 & -27.2 & 39 $\rightarrow$ 59 \\
 & OLMo 3 32B & -31.1 & -31.5 & -29.8 & 47 $\rightarrow$ 52 \\
 & OLMo 3 7B & -24.5 & -22.2 & -22.7 & 40 $\rightarrow$ 45 \\
 & Muse Glimmer 30B & -31.7 & -29.9 & -24.6 & 56 $\rightarrow$ 97 \\
\midrule
RTE & Qwen3.8-27B & -35.7 & -25.3 & -24.0 & 51 $\rightarrow$ 90 \\
 & OLMo 3 32B & -29.2 & -28.9 & -27.9 & 60 $\rightarrow$ 71 \\
 & OLMo 3 7B & -21.7 & -23.5 & -23.3 & 62 $\rightarrow$ 60 \\
 & Muse Glimmer 30B & -32.5 & -25.6 & -28.0 & 64 $\rightarrow$ 99 \\
 
\bottomrule
\end{tabular}}
\caption{Change (points) under \corr{drop-NOUN} in raw accuracy, centred accuracy and AUC, and the share of items (\%) given the most frequent answer, clean $\rightarrow$ perturbed. Zero-shot open LLMs.}
\label{tab:answer-bias}
\end{table}

\begin{table*}[h]
\centering
\small
\setlength{\tabcolsep}{4pt}
\begin{tabular}{lcccccc}
\toprule
Perturbation & Qwen3.8 & OLMo 32B & OLMo 7B & Glimmer & GPT-4o & DeepSeek \\
\midrule
\corr{drop-NOUN} & 63/29 & 70/19 & 68/14 & 47/25 & 56/26 & 49/25 \\
\corr{drop-VERB} & 74/60 & 79/49 & 76/50 & 66/56 & 71/64 & 64/58 \\
\corr{drop-ADJ} & 78/70 & 82/72 & 80/64 & 77/70 & 79/75 & 78/69 \\
\corr{drop-ADV} & 74/82 & 77/83 & 74/84 & 73/82 & --/-- & --/-- \\
\corr{drop-DET} & 92/93 & 92/91 & 93/93 & 91/93 & 91/95 & 92/95 \\
\corr{drop-PRON} & 97/99 & 97/97 & 97/92 & 94/97 & 99/97 & 96/99 \\
\corr{drop-NUM} & 97/68 & 98/71 & 97/65 & 96/70 & --/-- & --/-- \\
\corr{drop-CONJ} & 99/100 & 99/98 & 99/94 & 98/97 & --/-- & --/-- \\
\corr{drop-NOUN+PRON} & 60/28 & 67/17 & 66/13 & 43/23 & --/-- & --/-- \\
\corr{keep-NOUN+VERB} & 34/17 & 38/8 & 33/13 & 30/23 & 31/25 & 37/23 \\
\corr{keep-NOUN+VERB+ADJ} & 52/32 & 55/18 & 52/18 & 46/34 & --/-- & --/-- \\
\corr{keep-NOUN+ADV+VERB} & 54/22 & 59/9 & 57/14 & 49/28 & --/-- & --/-- \\
\corr{keep-NOUN+PRON+VERB} & 37/17 & 42/9 & 37/15 & 32/24 & --/-- & --/-- \\
\corr{keep-NOUN+VERB+ADV+ADJ} & 76/42 & 77/26 & 76/24 & 68/42 & --/-- & --/-- \\
\corr{shuffle-n1} & 46/15 & 52/2 & 52/2 & 32/18 & 46/26 & 42/21 \\
\corr{shuffle-n2} & 61/27 & 65/15 & 64/12 & 40/23 & --/-- & --/-- \\
\corr{shuffle-n3} & 69/34 & 72/20 & 70/19 & 50/32 & 64/44 & 55/33 \\
\corr{hyp-only} & 27/15 & 39/30 & 32/24 & 17/32 & 26/37 & 29/33 \\
\corr{swap} & 57/31 & 54/9 & 48/0 & 71/49 & --/-- & --/-- \\
 
\bottomrule
\end{tabular}
\caption{Share of above-chance performance retained ($R$, \%, from centred accuracy) on MNLI / mean of the three ANLI rounds, for every perturbation; hosted models were run on the core set only (--).}
\label{tab:full-llm}
\end{table*}

\begin{table}[h]
\centering
\small
\setlength{\tabcolsep}{3pt}
\resizebox{\columnwidth}{!}{%
\begin{tabular}{llccccc}
\toprule
Model & Task & Clean & No nouns & No verbs & Shuffled & Hyp.-only \\
\midrule
RoBERTa-base & CoLA & 58.8 & 15.3 & 4.4 & 1.7 & -- \\
 & MRPC & 88.1 & 79.6 & 84.2 & 75.2 & 31.6 \\
 & RTE & 70.8 & 54.0 & 67.6 & 57.2 & 48.0 \\
 & STS-B & 89.9 & 55.8 & 85.3 & 85.2 & 5.1 \\
\midrule
BERT-base & CoLA & 56.4 & 16.2 & 4.4 & 1.1 & -- \\
 & MRPC & 81.9 & 77.8 & 80.1 & 73.5 & 31.6 \\
 & RTE & 64.0 & 53.5 & 61.2 & 58.4 & 47.5 \\
 & STS-B & 88.4 & 56.6 & 84.2 & 86.1 & 3.6 \\
\midrule
ModernBERT-base & CoLA & 58.9 & 13.7 & 3.9 & 2.8 & -- \\
 & MRPC & 84.1 & 78.9 & 81.8 & 63.9 & 31.7 \\
 & RTE & 58.7 & 48.9 & 52.8 & 49.8 & 47.2 \\
 & STS-B & 89.2 & 53.1 & 84.2 & 81.7 & 1.3 \\
\midrule
XLM-R-base & CoLA & 43.5 & 19.2 & 5.0 & 2.4 & -- \\
 & MRPC & 84.2 & 77.9 & 80.0 & 76.1 & 31.5 \\
 & RTE & 59.2 & 51.1 & 59.5 & 54.1 & 51.1 \\
 & STS-B & 86.3 & 50.9 & 82.8 & 84.6 & 1.3 \\
 
\bottomrule
\end{tabular}}
\caption{Fine-tuned encoders trained on clean data (means over three seeds; GLUE metrics $\times100$).}
\label{tab:encoders}
\end{table}

\FloatBarrier
\section{Fresh Items}
\label{app:fresh}
Premises were taken from Wikipedia (CC BY-SA 4.0) and kept at most two per article; the infini-gram check dropped any premise with a 13-gram found in the OLMo~2 pre-training mix.
Generators alternated over premises, so each wrote half of the items in both styles.
Raters (14 volunteers, none a native English speaker, median 21 seconds per item) saw premise and hypothesis only, labelled each pair as entailment, neutral or contradiction from the premise alone, and could flag ambiguous, ungrammatical or knowledge-dependent pairs.
Most items received one label; on the 32 items with several, Krippendorff's $\alpha$ is 0.82, and the labels agree with the intended label on 85.6\% of the labelled generated items.
Revision pairs that differ only in case, punctuation or word order (28) were dropped.
On the kept items, the MNLI hypothesis-only classifier reaches 65.5\% on MNLI-style and 41.9\% on cue-free items (54.3\% on MNLI itself), and 39\% of MNLI-style contradictions contain a negation word, against none of the cue-free ones.
The items remain private until publication so that they stay unseen.

\paragraph{Annotation guidelines.} Raters saw the following instructions verbatim:
\begin{quote}\small
For each pair, choose a label judged only from the premise as written:
Entailment (e): if the premise is true, the hypothesis must be true.
Neutral (n): the hypothesis may or may not be true.
Contradiction (c): if the premise is true, the hypothesis must be false.
Near-identical sentence pairs can just get e.
Under Problem, note anything that makes the item unusable (ambiguous, ungrammatical, needs outside knowledge). Leave it empty if the item is fine.
Label independently and keep the items private: don't share them or paste them anywhere online.
Do as many as you like and stop whenever you want.
\end{quote}
On each item, raters could tick \emph{ambiguous}, \emph{ungrammatical} or \emph{needs outside knowledge} and add free text before choosing a label.

\FloatBarrier
\section{Compute and Cost}
Fine-tuning, log-probability scoring and the contamination checks ran on NVIDIA GH200 nodes, about 59 GPU-hours in total.
The hosted models were queried through Azure AI Foundry for USD 126, including the generation of the fresh hypotheses.

\FloatBarrier
\section{Contamination Indicators}
\label{app:contamination}
Table~\ref{tab:contamination} gives the three contamination indicators discussed in Section~\ref{sec:contamination}.

\begin{table}[t]
\centering
\small
\setlength{\tabcolsep}{3.5pt}
\resizebox{\columnwidth}{!}{%
\begin{tabular}{lcccccc}
\toprule
 & \multicolumn{2}{c}{TS-guessing} & \multicolumn{2}{c}{Min-K\%++} & \multicolumn{2}{c}{infini-gram} \\
\cmidrule(lr){2-3}\cmidrule(lr){4-5}\cmidrule(lr){6-7}
Task & copy & no copy & Qwen & OLMo & Prem. & Hyp. \\
\midrule
MNLI & 56.3 & 29.0 & -1.89 & -1.88 & 21 & 18/448 \\
RTE & 66.4 & 32.4 & -1.78 & -1.49 & 24 & 3/25 \\
ANLI R1 & 67.5 & 34.0 & -1.79 & -1.07 & 72 & 10/273 \\
ANLI R2 & 69.2 & 31.3 & -1.84 & -1.03 & 71 & 5/218 \\
ANLI R3 & 59.2 & 26.6 & -2.18 & -1.49 & 34 & 3/228 \\
QNLI & 42.6 & 34.6 & -3.05 & -1.70 & -- & -- \\
MRPC & 70.1 & 27.5 & -1.87 & -1.69 & -- & -- \\
SST-2 & 13.8 & 13.6 & -2.17 & -1.80 & -- & -- \\
 
\bottomrule
\end{tabular}}
\caption{Contamination indicators. TS-guessing: exact-match rate (\%) of Qwen3.8-27B filling in a masked word, when the word may also occur in the other field (copy) and when it may not. Min-K\%++: mean score (higher suggests seen). infini-gram: premises (or first sentences) found in OLMo's pre-training data (\%), and hypotheses of at least 13 words found / all such hypotheses.}
\label{tab:contamination}
\end{table}

\FloatBarrier
\section{BioASQ}
\label{app:bioasq}
\begin{table}[h]
\centering
\small
\setlength{\tabcolsep}{3pt}
\resizebox{\columnwidth}{!}{%
\begin{tabular}{lcccccc}
\toprule
& \multicolumn{2}{c}{No evidence} & \multicolumn{2}{c}{No nouns} & \multicolumn{2}{c}{Nouns+verbs} \\
\cmidrule(lr){2-3}\cmidrule(lr){4-5}\cmidrule(lr){6-7}
Model & MNLI & BioASQ & MNLI & BioASQ & MNLI & BioASQ \\
\midrule
Qwen3.8-27B & 26 & 83 & 72 & 97 & 44 & 95 \\
OLMo 3 32B & 51 & 77 & 79 & 94 & 47 & 89 \\
OLMo 3 7B & 41 & 64 & 78 & 88 & 43 & 83 \\
GPT-4o & 27 & 89 & 60 & 98 & 38 & 92 \\
DeepSeek-V4-Pro & 29 & 88 & 55 & 98 & 40 & 96 \\
 
\bottomrule
\end{tabular}}
\caption{Retention rate of above-chance AUC ($R$, \%) without evidence (MNLI: hypothesis only, BioASQ: question only) and under \corr{drop-NOUN} and \corr{keep-NOUN+VERB} for MNLI and BioASQ yes/no questions. Since models respond \emph{yes} by default when BioASQ items are degraded, AUC was used for both datasets.}
\label{tab:bioasq}
\end{table}

We use the yes/no questions of the BioASQ Task~b training set 11b (1,271) and of the test sets of 12b, 13b and 14b (248), with their gold snippets; the perturbations are applied to the snippets only.
Answers are scored by the log-probabilities of \emph{yes} and \emph{no}, computed by us for the open models and returned by the provider for the hosted models.
One item blocked by a provider's content filter is dropped for the hosted models.
Llama-3.3-70B, evaluated on BioASQ only, retains 95\% of its above-chance AUC under \corr{drop-NOUN}.
Measured from the question-only baseline instead of chance, the snippets keep 51--73\% of their contribution under \corr{keep-NOUN+VERB} for the open models and DeepSeek-V4-Pro, against $-8$ to 25\% on MNLI measured from the hypothesis-only baseline; for GPT-4o the figure is 23\% (95\% interval $-14$ to 50), which does not separate it from MNLI (15\%), since its clean and question-only AUC differ by only five points.

\paragraph{Question cues.}
A logistic regression on question uni- and bigrams reaches a macro F1 of 0.61 on the training set (five-fold cross-validation) but 64.5\% accuracy on the pooled test sets, below the majority baseline of 64.9\%.
Splitting the gold-\emph{no} items by this classifier, as for MNLI, gives intervals above zero in only 5 of 24 model--perturbation cells, and no reversal under \corr{keep-NOUN+VERB}.
We restrict the split to gold-\emph{no} items because over all items it is confounded with the models' default answer \emph{yes}.

\paragraph{Answer default.}
Almost all of the loss falls on gold-\emph{no} items, which the models increasingly answer with \emph{yes} as the evidence degrades; gold-\emph{yes} accuracy stays at 97--100\% under every perturbation.
Qwen3.8 is the exception, losing on both classes.
AUC, which does not depend on a decision threshold, corrects for this default.

\paragraph{Memorisation.}
By the developers' stated training cut-offs, the 12b--14b questions postdate the training data of GPT-4o and Llama-3.3-70B, 13b and 14b that of OLMo~3, and, by a third-party report, 14b that of DeepSeek-V4-Pro; Qwen3.8 has no published cut-off.
On factoid questions, the share of performance kept without the evidence on these newer questions overlaps that on the training set for every model; yes/no questions are too few per year for this comparison.
For the 1,668 factoid questions, we compared items most of whose snippets contain a 13-gram found in the OLMo~2 pre-training mix with the rest, stratified by the number of snippets; the difference in the drop without evidence is within its interval for every model in the pooled data.

\end{document}